\documentclass{article}

\usepackage{microtype}
\usepackage{graphicx}
\usepackage{subfigure}
\usepackage{booktabs} 

\usepackage{hyperref}

\usepackage[accepted]{icml2024}

\usepackage{amsmath}
\usepackage{amssymb}
\usepackage{mathtools}
\usepackage{amsthm}

\usepackage[capitalize,noabbrev]{cleveref}

\theoremstyle{plain}

\theoremstyle{definition}

\theoremstyle{remark}

\usepackage[textsize=tiny]{todonotes}

\icmltitlerunning{A Geospatial Approach to Predicting Desert Locust Breeding Grounds in Africa}

\begin{document}

\twocolumn[
\icmltitle{A Geospatial Approach to Predicting \\ Desert Locust Breeding Grounds in Africa}



\icmlsetsymbol{equal}{*}

\begin{icmlauthorlist}
\icmlauthor{Ibrahim Salihu Yusuf}{comp}
\icmlauthor{Mukhtar Opeyemi Yusuf}{comp}
\icmlauthor{Kobby Panford-Quainoo}{comp}
\icmlauthor{Arnu Pretorius}{comp}
\end{icmlauthorlist}

\icmlaffiliation{comp}{InstaDeep}

\icmlcorrespondingauthor{Ibrahim Salihu Yusuf}{i.yusuf@instadeep.com}

\icmlkeywords{Machine Learning, Geospatial ML, Remote Sensing}

\vskip 0.3in
]



\printAffiliationsAndNotice{}  

\begin{abstract}
Desert locust swarms present a major threat to agriculture and food security. Addressing this challenge, our study develops an operationally-ready model for predicting locust breeding grounds, which has the potential to enhance early warning systems and targeted control measures. We curated a dataset from the United Nations Food and Agriculture Organization's (UN-FAO) locust observation records and analyzed it using two types of spatio-temporal input features: remotely-sensed environmental and climate data as well as multi-spectral earth observation images. Our approach employed custom deep learning models (three-dimensional and LSTM-based recurrent convolutional networks), along with the geospatial foundational model Prithvi recently released by \cite{jakubik2023foundation}. These models notably outperformed existing baselines, with the Prithvi-based model, fine-tuned on multi-spectral images from NASA’s Harmonized Landsat and Sentinel-2 (HLS) dataset, achieving the highest accuracy, F1 and ROC-AUC scores (83.03\%, 81.53\% and 87.69\%, respectively). A significant finding from our research is that multi-spectral earth observation images alone are sufficient for effective locust breeding ground prediction without the need to explicitly incorporate climatic or environmental features. 
\end{abstract}



\section{Introduction}
\label{introduction}

Desert Locusts (DL) are voracious migratory pests that can form large swarms, causing significant damage to crops and leading to food crises that affect humans and livestock \cite{peng2020review}. Over the years, DL have posed a significant threat to food security in Africa. Their ability to swiftly migrate over long distances and across geographical boundaries, makes it extremely difficult to coordinate preventative measures between control teams. Although various attempts have been made to mitigate the DL threat \cite{enns2022disaster}, much remains to be done. 

DL exhibit phase polymorphism (also called polyphenism) which causes them to change from a solitarious to a gregarious phase and vice versa. Solitarious DL are shy, sedentary and do not move much, and with their green or brownish cryptic colour hide during the day \citep{Pflger2021OneHY}. They also avoid other locusts, except for mating, and are reported to migrate at night \citep{uvarov1977grasshoppers,Pflger2021OneHY}. In contrast, gregarious DL are conspicuous and reveal anti-predator warning colours in bright yellow and black \cite{Pflger2021OneHY}. Gregarious locusts are very active, and they aggregate both as nymphs (marching hopper bands) and adults (swarms), and they fly by day and roost overnight in trees \cite{Pflger2021OneHY}. It has been observed that a major stimulus for polyphenism in solitarious DL is crowding. When solitarious DL are crowded, regularly touching the hind femur of one another, or perceiving the smell and sight of others, this causes a solitarious individual to become gregarious \cite{rogers2003mechanosensory, anstey2009serotonin}. When adult locusts become gregarious they form large swarms; a swarm measuring a single square kilometre can contain up to 80 million individuals (or more) that can fly up to 90 miles a day and consume an equivalent of their body mass in green vegetation every day \citep{uvarov1957aridity}. This is equivalent to the food that would be consumed by 35,000 people. As they migrate in search of food, each female lays up to 80 eggs in a pod. If the environmental conditions are suitable for breeding, the population of DL can grow exponentially within a short period of time, thereby resulting in a plague. 

Historically, DL plagues have caused significant harm to both human and animal lives \cite{GROSS2021R459, agronomy13030819}. Various initiatives led by the UN-FAO and other stakeholders have aimed to mitigate this threat. Despite these efforts, there remain opportunities to enhance the effectiveness of measures against the adverse impact of DL on both animal and crop production. Given the nature of their lifecycle, whereby adults lay eggs that hatch in about 2 weeks and grow to become adults that form swarms in about 4 months, there is a need to develop a comprehensive solution to mitigating locust plagues. The first stage in locusts' lifecycle is oviposition by adult females which later results in hoppers that have limited locomotion abilities \citep{symmons2001desert}. 

During this phase, known as the breeding stage, directing control efforts towards eradicating locust breeding appears to be a more effective approach. If successful at this stage, it means that some future generations of locusts have been eliminated and, if consistently carried out, after some years it is possible to reduce the negative impact of locust plagues to a bare minimum. This approach would first involve the identification of actual locust breeding grounds and subsequently the application of an effective control operation such as pesticide spraying.

In this paper, we present a set of new custom deep learning architectures as well as a fine-tuned foundational model specifically for DL breeding ground prediction. These models were created by first curating a breeding dataset from locust observation data collected by the UN-FAO and then exploring new modelling strategies using various remotely-sensed features and multi-spectral earth observation images. Whilst past research aimed at identifying regions \emph{favourable} to breeding, our focus is more specific. We care about identifying regions where locusts would be found \emph{present}, copulating and/or laying eggs, i.e.\ \textit{actual} breeding grounds across Africa. However, even with this more specific task, we show that our models significantly outperform existing baselines,  achieving accuracy, F1 and ROC-AUC scores of 83.03\%, 81.53\% and 87.69\%, respectively. By openly releasing our models, we hope this work might assist locust control agencies and potentially improve early warning systems and the effectiveness of targeted control measures.



\subsection{Related Work}
\label{related_work}
Several researchers have approached locust breeding ground prediction using machine learning \cite{gomez2018machine, kimathi2020prediction, gomez2021prediction, KLEIN2022102672}. Here we give a brief overview of this work.

Machine learning has been instrumental in assessing the role of soil moisture (SM) in predicting DL breeding grounds. \cite{gomez2018machine} used a machine learning approach to predict DL breeding grounds based on SM data from European Space Agency Climate Change Initiative (ESA-CCI) \footnote{https://esa-soilmoisture-cci.org/}. In their study area of Mauritania, and period of 1985-2015, random forest achieves the best performance in evaluating the link between SM and DL hoppers' presence with Kappa and ROC-AUC scores of 0.95 and 0.74 respectively. Their study shows that ESA-CCI SM was a significant predictor of DL breeding grounds in areas around Mauritania. They also demonstrate that a location is deemed favourable for breeding when the SM minimum value is over 0.07 $m^3/m^3$ within 6 days or more. To improve their work, \cite{gomez2021prediction} addressed the lack of automated and operational procedures for predicting DL breeding grounds in near real-time (NRT). They indicated that the ESA-CCI SM data is released with several months of lag. Therefore, they used SM data from the Soil Moisture and Ocean Salinity (SMOS) satellite data product\footnote{https://earth.esa.int/eogateway/catalog/smos-nrt-data-products}, which updates three times every month, to predict DL breeding grounds. Their study area covers the entire DL recession area between 2016–2018, which includes more than 30 countries from West Africa to West India, spanning about 16 million square kilometers. They evaluated six machine learning models and found a Weighted k-Nearest Neighbors approach to have the best performance with a Kappa and ROC-AUC score of 0.50, and 0.80 respectively.

Some studies have combined SM data along with other bio-climatic data to improve model generalization. \cite{kimathi2020prediction} is one of these studies that utilizes machine learning algorithms to predict potential DL breeding grounds in East Africa. They considered key bio-climatic factors such as temperature and rainfall, as well as edaphic factors such as sand and moisture contents. Using the MaxEnt algorithm, they trained models on Morocco, Mauritania and Saudi Arabia and found Morocco model parameters to have the best generalization with an AUC score of 0.82. The findings of this study revealed that vast areas of Kenya and Sudan, along with the northeastern regions of Uganda, and the southeastern and northern regions of South Sudan, were at a high risk of providing a suitable breeding environment for DL in the period between February and April 2020.

So far, the data considered in all of these studies are temporal and vary rapidly over time. Some studies have incorporated certain static ecological data that can be useful for DL breeding survival. \cite{KLEIN2022102672} introduce a fused multi-scale approach for predicting suitable breeding habitat for different locust species in different areas, including \textit{Calliptamus italicus, CIT} (Italian locust) in Pavlodar oblast (Kazakhstan), \textit{Dociostaurus maroccanus, DMA} (Moroccan locust) in Turkistan oblast (Kazakhstan), and \textit{Schistocerca gregaria} (Desert Locust) in the Awash river basin (Ethiopia, Djibouti, Somalia). They incorporated up-to-date land surface parameters, vegetation development, and other relevant environmental factors into their model, along with climatic and soil preferences derived from ecological niche modelling (ENM). They emphasize the importance of considering actual changes in the landscape and human interactions for understanding locust outbreaks and defining suitable breeding grounds. To address this, the authors propose incorporating variables obtained from Sentinel-2\footnote{https://sentinel.esa.int/web/sentinel/missions/sentinel-2} (high-resolution remote sensing data) time-series analysis to describe the current state of the land, thereby refining the suitable breeding grounds within the model. For the year 2019, the model was validated using field observations and achieved an AUC performance of 0.747 for CIT, 0.850 for DMA and 0.801 for DL.

Although classical machine learning models have shown promise for locust control, their ability to handle complex temporal data is limited. In contrast, certain deep learning models are well suited for such data. As an example, \cite{tabar2021plan} introduced the Predictor of Locust Activity and movemeNt (PLAN) model, a deep learning approach to forecasting DL migration patterns by integrating crowdsourced observations (data from PlantVillage\footnote{https://plantvillage.psu.edu/}) and remote-sensed data. Inputs to their model are categorized into three categories: (1) static variables (actual evapotranspiration, sand content, total biomass productivity and elevation), (2) temporal variables (soil moisture, precipitation and wind speed) and (3) historical locust observations. Each category is processed by a different sub-component of the model and later fused together for the final prediction. Their model achieved an AUC score of 0.89 in forecasting the presence of DL.

Previous studies on desert locust breeding ground prediction have considered different datasets, input variables and study areas, making it difficult to benchmark the performance of the model from each study. The remotely-sensed variables used are also from different satellite data products with varying spatial and temporal resolutions as well as update frequency. These compounding issues make it difficult to build a robust and operationally-ready model that can be used by regional governments and control agencies such as the UN-FAO for administering control activities towards eliminating the threat posed by DL. In this study, our goal is to build such a robust and operationally-ready model for predicting DL breeding grounds.

\section{Methods}
\label{exp:methods}

Locust breeding ground prediction is framed as a binary classification task. We define a set of geographical locations as \(\mathcal{L}\), where each location \(l\) has a specific coordinate. Each location \(l\) has an associated binary label \(y_l\), indicating if it's an actual breeding ground (\(y_l = 1\)) or not (\(y_l = 0\)). For every location, a feature vector \(x_l\) is introduced to represent spatio-temporal data. This vector incorporates components including temporal variables and non-temporal variables. 

In essence, the goal is to learn a function \(f: \mathcal{X} \rightarrow \{0, 1\}\) based on labeled data \(\{(x_l, y_l)\}\), predicting the breeding ground label for any feature vector \(x_l\). The subsequent sections detail the data preparation for this problem.

\subsection{Data}
We present our data sources, collection methods, and the final data processing used in our models.

\subsubsection{Locust Observation Records.}
To learn the behaviour of locusts that would help in identifying breeding grounds, there is a need to have enough locust observation data. While there are locust observation data collected by regional authorities in some countries, they are limited in their spatial and temporal coverage and accessibility. The primary source of locust observation data that overcomes the aforementioned limitations is the UN-FAO Locust Hub\footnote{https://locust-hub-hqfao.hub.arcgis.com}. The UN-FAO has been collecting locust observation data over the past 48 years (1975-Present) spanning different locust stages (Hoppers, Bands, Adults, and Swarms) across Africa and Asia. The data contains geolocation records and other environmental conditions of the observed site and is made publicly available via the Locust Hub. The data is collected based on guidelines recommended by the UN-FAO \citep{2001desert} to be used during field surveys.

\subsubsection{Breeding Data.}
\label{sec:breeding_data}
The locust observation data from the FAO describes DL observed at different stages (Hoppers, Bands, Adults, and Swarms). For our use case, we are interested in the records that describe the breeding stage. We curate the appropriate data for this stage in a manner similar to \cite{KLEIN2022102672} by including records of adults found laying eggs and those of early-stage instars (insect development stage), which also depict successful incubation and nymph hatching \citep{KLEIN2022102672}. 

\begin{figure}
\centering
\includegraphics[trim={0 1.5cm 0 0}, width=1.0\linewidth, clip]{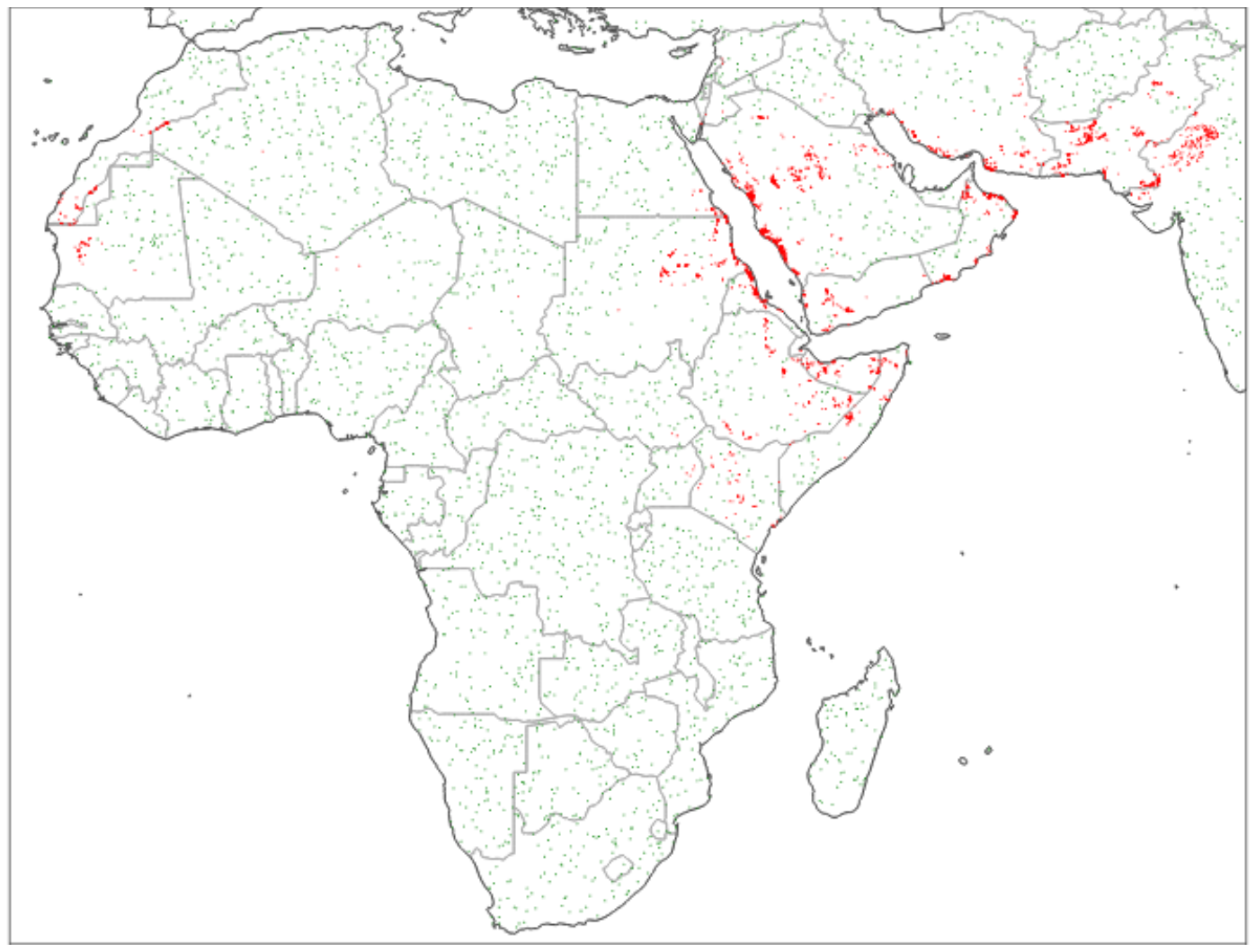}
\caption{ \textit{Visualization of Breeding Data.} Green and Red points represent non-breeding and breeding locations respectively. There is an equal number of green and red points. While breeding locations are concentrated, non-breeding locations are spread out due to random sampling during pseudo-absence generation.}
\label{breeding_data}
\end{figure}

In the domain of species distribution modeling, data collected from the field are mostly presence records. However, the UN-FAO data includes observations of sites where locusts were expected to be found (favorable locations) but were missing. We wanted to consider these records as a subset of the absence records, but from our preliminary experiments, we found them to be biased. As a result, similar to \cite{gomez2021prediction}, we discarded them and opted to generate pseudo-absence records as prevalent in previous studies \cite{gomez2018machine, gomez2021prediction, KLEIN2022102672}. We perform pseudo-absence generation using a random sampling technique \citep{ITURBIDE2015166, yusuf2022pseudoabsence} while maintaining a buffer zone around each presence observation. A combination of presence and pseudo-absence records, described in table \ref{table:breeding_records}, makes up a robust dataset for learning to classify actual locust breeding grounds. Figure \ref{breeding_data} also shows a visual representation of our curated breeding data.

\begin{table}
\caption{Description of Curated Breeding Records}
\label{table:breeding_records}
\centering
\resizebox{0.9\columnwidth}{!}{\begin{tabular}{l|ccc}
\toprule
\textbf{Split} & \textbf{Non-Breeding (0)} & \textbf{Breeding (1)} & \textbf{Date Range} \\
\midrule
Train & 2238 & 2238 & 2020-01-01 to 2021-04-21 \\
Validation & 154 & 154 & 2021-04-22 to 2021-07-09 \\
Test & 820 & 820 & 2021-07-10 to 2023-07-30 \\
\bottomrule
\end{tabular}}
\end{table}

\subsection{Input Features. }
\label{sec:input_features}
Using the breeding data created in \ref{sec:breeding_data}, which contains only geo\-location and time information, we derive input features from remotely-sensed environmental and climatic variables as well as earth observation data. In each case, we prioritize satellite data products with higher spatial-temporal resolution and update frequency. 

\subsubsection{Remotely-Sensed Variables}
\label{sec:rs_input_features}
Similar to previous studies, we sourced environmental and climatic features that are suitable for locust breeding. These features are either static or temporal. We drew inspiration from \cite{tabar2021plan} and crafted them into a spatio-temporal data representation. By dividing the earth's surface into a grid of 0.1-degree resolution, we extracted an $n \times n$ grid centered on the observed location, where $n$ is odd with a default value of 7. Two data feature groups were prepared:
\begin{itemize}
    \item[1] \textbf{Temporal}: Incorporating soil moisture, precipitation, and fraction of vegetation cover, we constructed a spatio-temporal representation. For each record in our data, we retrieved a 96-day historical record of each temporal variable and resampled the data by computing a mean for each 3-day period. Following the resampling process, we obtained 30 time-steps for each variable, resulting in an input of shape $30 \times 7 \times 7 \times 3$.
    
    \item[2] \textbf{Non-temporal}: Combining all 14 variables from the TerraClimate \footnote{https://www.climatologylab.org/terraclimate.html} with SoilGrid's \textit{sand\_5-15cm\_mean} \footnote{https://www.isric.org/explore/soilgrids}, Copernicus \textit{land\_cover\_2019} \footnote{https://land.copernicus.eu/global/products/lc} and ALOS WORLD 3D - 30m \textit{wadis} \footnote{https://www.eorc.jaxa.jp/ALOS/en/dataset/aw3d30/aw3d30\_e.htm} data products, we constructed a matrix of $7 \times 7 \times 17$.
\end{itemize}

\begin{figure}
\centering
\includegraphics[trim={0 4cm 0 0}, width=0.95\linewidth, clip]{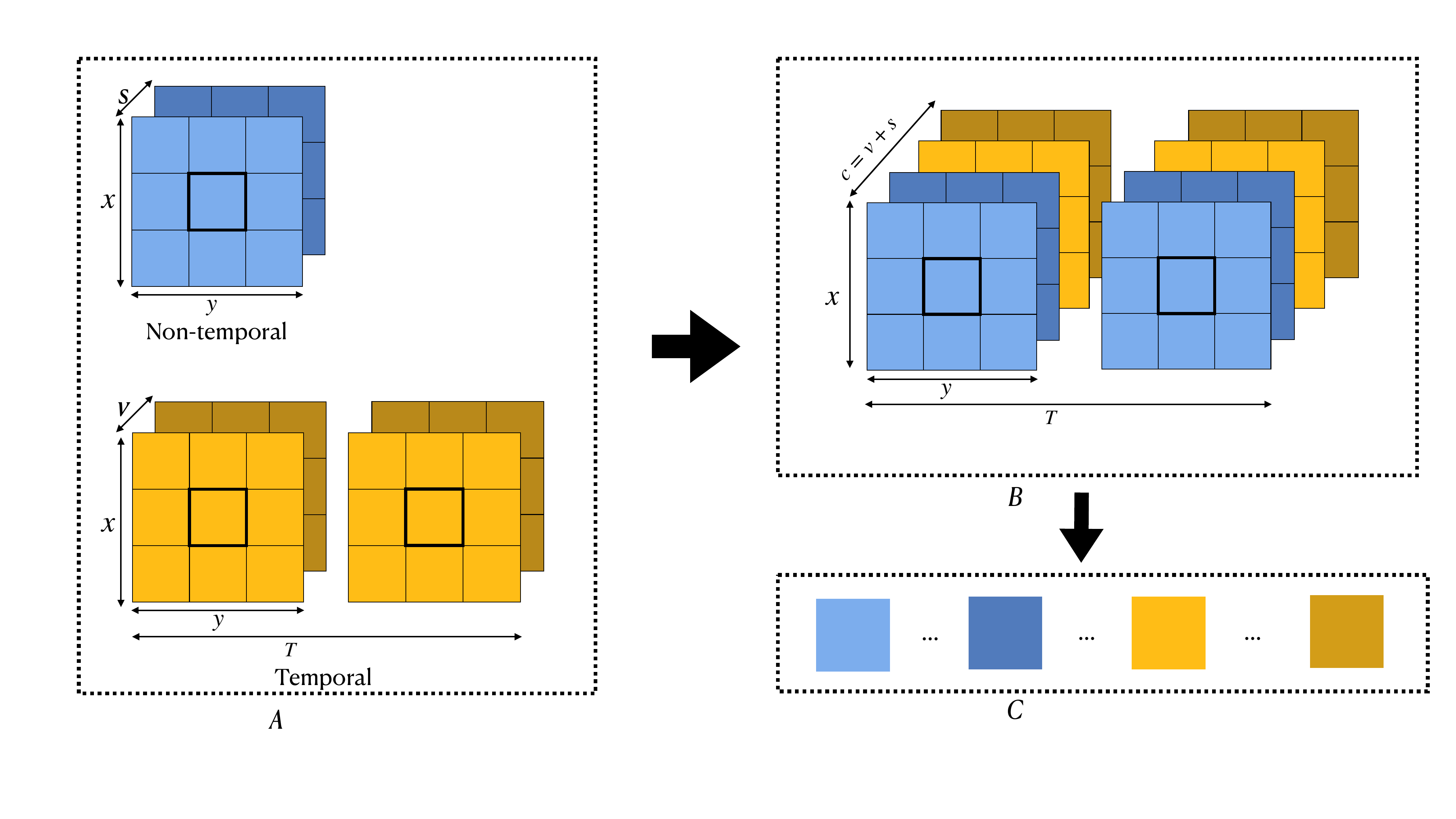}
\caption{\textit{Visual Representation of Spatiotemporal Input Features.} The spatial dimensions are denoted by $x$ and $y$, while $T$ represents the temporal dimension. $v$ and $s$ indicate the number of temporal and non-temporal variables, respectively. (A) corresponds to the input fed into our PLAN-LB model. (B) shows the spatiotemporal representation utilized by our Conv3D and ConvLSTM models. Lastly, (C) presents the input employed for SVM and logistic regression models.}
\label{data_repr}
\end{figure}

Figure \ref{data_repr} visually outlines these feature groups. Here $x$, and $y$ denote spatial dimensions, and $T$ is the temporal dimension (the number of time-steps after resampling). $v$ and $s$ are the number of temporal, and non-temporal features, respectively. The individual representation in (A) is fed to our PLAN-based model while the spatio-temporal representation in (B) is fed to the other deep learning models that utilize remotely-sensed input features. For our classical models, we flatten each data group and concatenate them before feeding the input to the model, as illustrated in (C).

\begin{figure*}
    \centering
    \includegraphics[width=0.7\linewidth, trim={0 7cm 10cm 0}, clip]{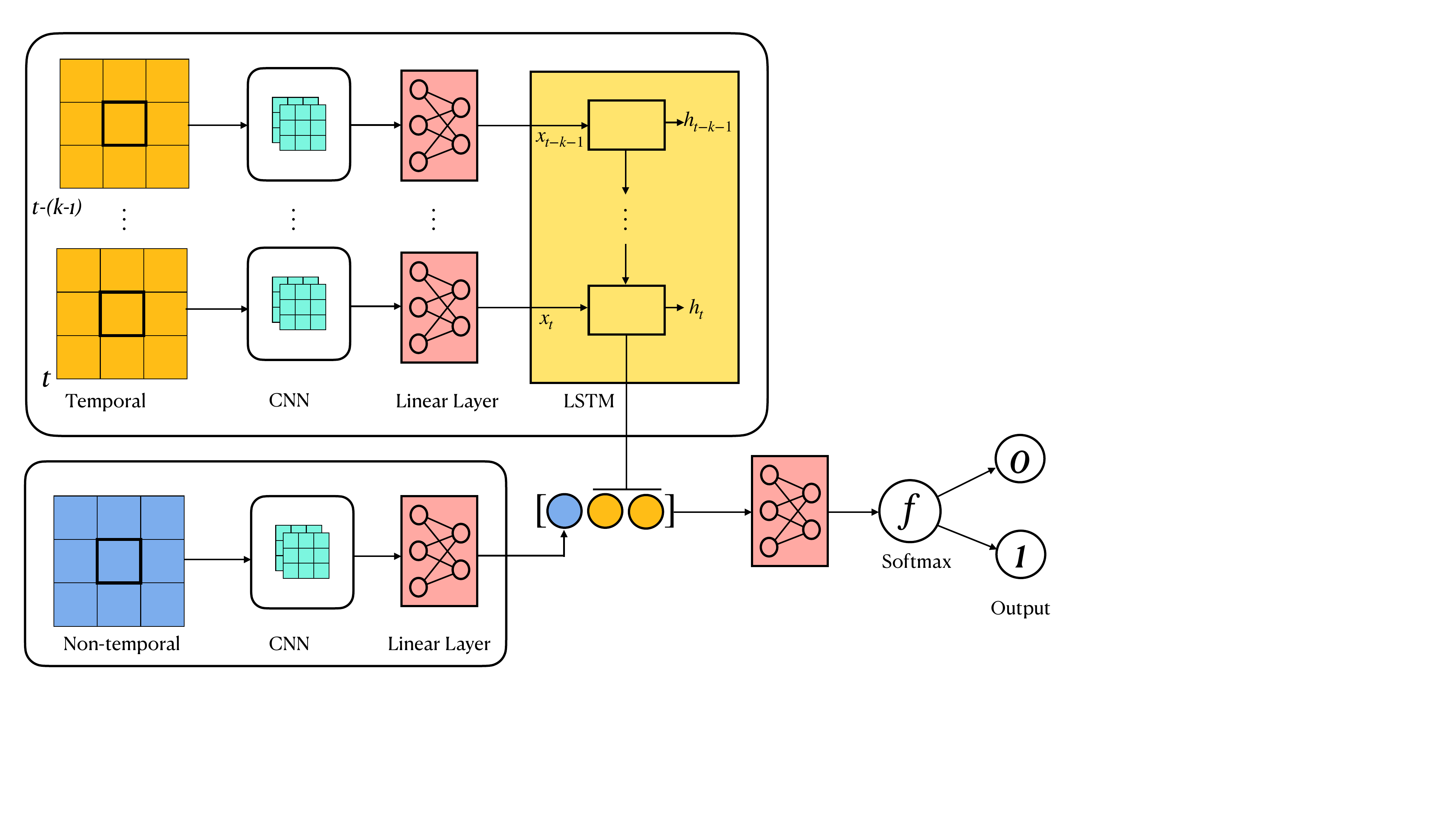}
    \hspace{-3cm}
    \caption{\textit{PLAN-LB Model Architecture. } This model was derived from PLAN's model and it independently processes the temporal and non-temporal inputs. The temporal module encodes each entry in the temporal series into a feature vector. Subsequently, the series of resulting feature vectors undergo processing via an LSTM block in a many-to-one configuration. The non-temporal module similarly encodes the non-temporal input into a feature vector. The outputs from both modules are then concatenated and forwarded to a final linear layer for classification}
    \label{fig:plan_model}
\end{figure*}

\subsubsection{Earth Observation Data}
\label{sec:hls_input_features}
Unlike remotely sensed features such as soil moisture, elevation etc. earth observation data are multi-spectral images of the earth's surface. These images have spectral bands ranging from visible light to the near-infrared (NIR) and shortwave infrared (SWIR) part of the electromagnetic spectrum. This enables detailed observations of vegetation, soil and water cover, inland waterways, and coastal areas. We hypothesize that the information contained in these spectral images can serve the same purpose as remotely-sensed features in detecting desert locust breeding grounds. 

NASA's Harmonized Landsat and Sentinel-2 (HLS) \cite{CLAVERIE2018145} \footnote{https://hls.gsfc.nasa.gov/} dataset, which provides a high temporal (2-3 days) and spatial (30m) multi-spectral image of the earth's surface makes a suitable source of input features for an operational locust breeding ground prediction model. Using the HLS satellite data and our curated breeding records, we create spatio-temporal ``chips'' (subsection of a larger geospatial image) of size $224 \times 224$ for locust breeding ground prediction which comprise 3 temporal steps with a size of 30 days. Each step also includes 6 multi-spectral bands (Blue, Green, Red, Narrow NIR, SWIR 1, SWIR 2). This implies that the resulting input features of shape $3 \times 6 \times 224 \times 224$ represent an observation of a specific area over the past 90 days.

\begin{figure*}
    \centering
    \includegraphics[trim={0 6cm 0 0}, width=0.75\linewidth, clip]{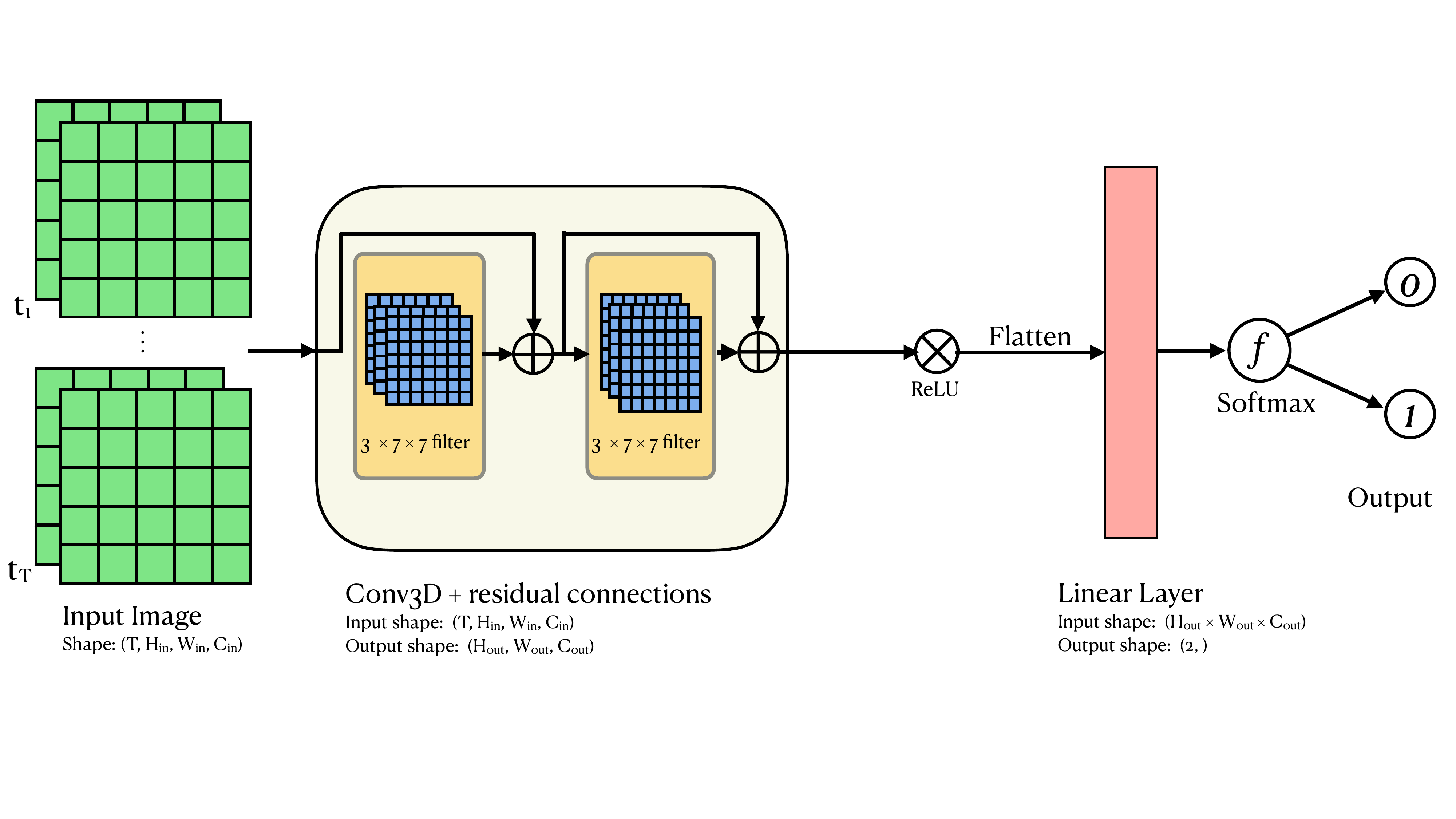}
    \caption{\textit{Conv3D Model Architecture. } Our Conv3D model features two residual layers with Conv3D blocks, layer normalization, and ReLU activation. It employs a kernel size of $(3, 7, 7)$ and retains spatial dimensions using a ``same'' padding approach. The model concludes with average pooling and a softmax activation output layer, producing a probability score for locust breeding likelihood. Given that the input features pertain to a specific point location, the output provides a classification indicating whether the point is a breeding or non-breeding ground.}
    \label{fig:model_conv3d}
\end{figure*}

\subsection{Models}
\label{models}

In this section, we present deep learning models designed to model the complex relationships between the various spatio-temporal features from the two categories of data discussed in the previous section. By leveraging deep learning, our models aim to uncover intricate patterns and dependencies within the input features that will aid in locust breeding ground prediction. In the following subsections, we describe the architecture and functioning of these models.


\subsubsection{PLAN for Locust Breeding (PLAN-LB)}
This model is a modified version of the PLAN model \cite{tabar2021plan}. Since our input features do not include historical locust observations, as seen in PLAN, we employ only two modules for processing the temporal and non-temporal input. As explained in Section \ref{sec:rs_input_features}, our inputs have spatial dimensions of $7 \times 7$ to incorporate neighboring information, in contrast to the point values utilized in PLAN. Consequently, the two modules we utilized are convolutional modules, as depicted in Figure \ref{fig:plan_model}. The temporal module independently encodes each entry in the temporal series into a feature vector. Subsequently, the series of resulting feature vectors undergo processing via an LSTM block in a many-to-one configuration. The non-temporal module similarly encodes the non-temporal input into a feature vector. The outputs from both modules are then concatenated and forwarded to a final linear layer for classification.
    
\subsubsection{3D convolutional network (Conv3D)} 

This model learns from the spatio-temporal input features using a three-dimensional convolutional network architecture. Conv3D has been effective in domains like action recognition, medical analysis, and geospatial analysis \citep{wang2018discrimination, wu20183d, lee3d, liu2019spatiotemporal, lee2021efficient, duan2022revisiting}. Our Conv3D model features two residual layers with Conv3D blocks, layer normalization, and ReLU activation. It employs a kernel size of $(3, 7, 7)$ and retains spatial dimensions using a ``same'' padding approach. The model concludes with average pooling and a softmax activation output layer, producing a probability score for locust breeding likelihood. The full architecture is displayed in Figure \ref{fig:model_conv3d}. 
    
\subsubsection{Convolutional LSTM (ConvLSTM)}
\begin{figure*}
    \centering
    \includegraphics[width=0.75\linewidth, trim={0 4.5cm 0 0}, clip]{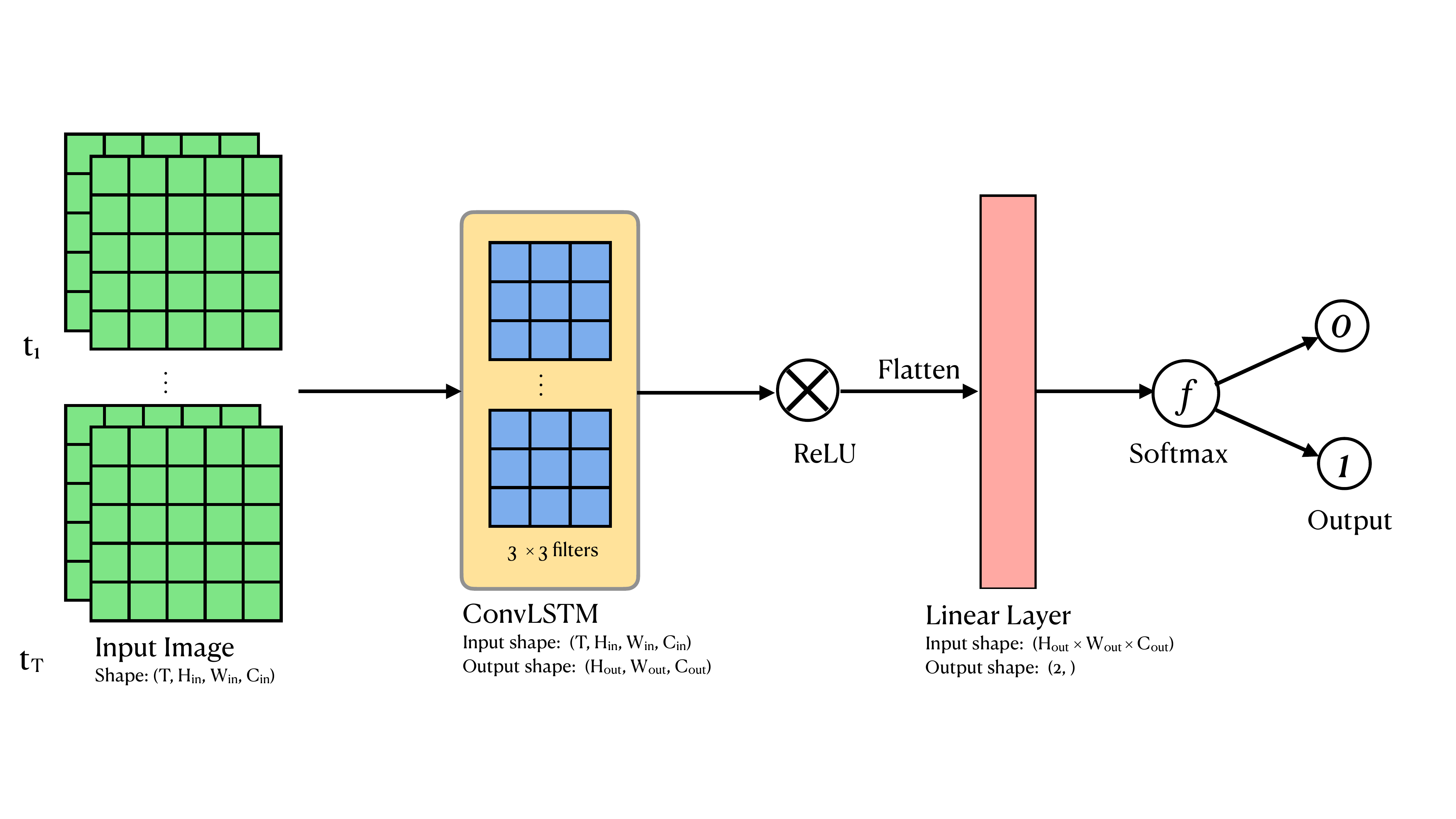}
    \caption{\textit{ConvLSTM Model Architecture. } Our ConvLSTM uses a kernel size of $(3, 3)$ for convolutional operations, followed by a ReLU activation. A linear layer then transforms its outputs, and a softmax activation provides a probability score for locust breeding likelihood. Given that the input features pertain to a specific point location, the output provides a classification indicating whether the point is a breeding or non-breeding ground.}
    \label{fig:model_convlstm}
\end{figure*}

A convolutional variant of the standard LSTM (Long Short-Term Memory) \citep{hochreiter1997long} recurrent network, ConvLSTM incorporates convolutional operations within the recurrent structure, both in the input-to-state and state-to-state transitions \citep{shi2015convolutional}. It is specifically designed to capture spatio-temporal dependencies in data and has been effective in applications such as video analysis and weather forecasting \citep{shi2015convolutional, sanchez2020exploiting, moishin2021designing}. Our ConvLSTM uses a kernel size of $(3, 3)$ for convolutional operations, followed by a ReLU activation. A linear layer then transforms its outputs, and a softmax activation provides a probability score for locust breeding likelihood. The architecture is visualized in Figure \ref{fig:model_convlstm}.

\subsubsection{Prithvi for Locust Breeding (Prithvi-LB)}
\begin{figure}[!h]
    \centering
    \includegraphics[width=0.95\linewidth]{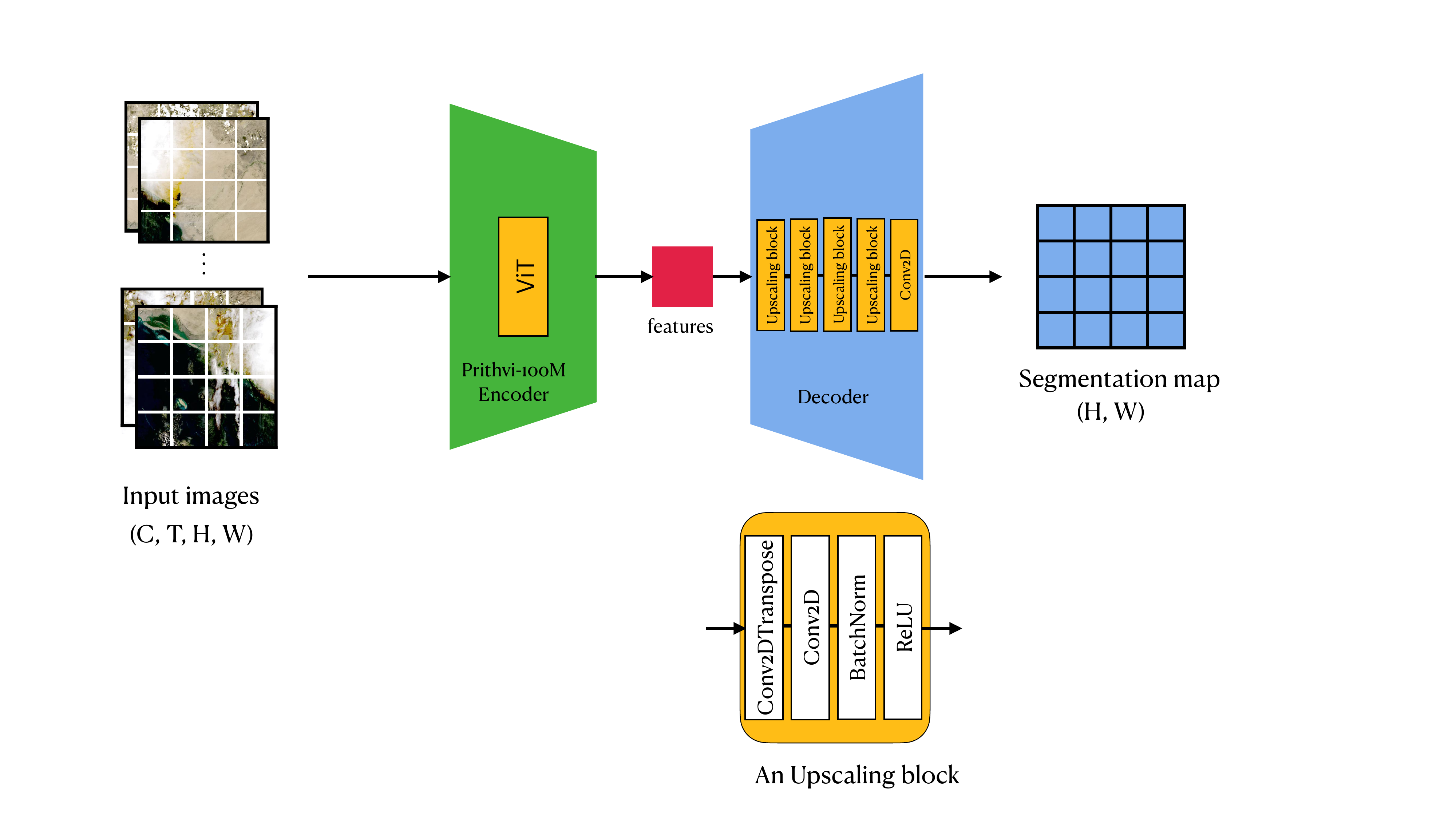}
    \caption{\textit{Prithvi-LB Model Architecture. } This custom model was derived by adding a custom decoder layer atop the pre-trained Prithvi vision transformer encoder. The custom decoder consist of a stack of upsampling blocks followed by a final two-dimensional convolutional layer that produces the output segmentation map. The segmentation map classifies each 30-square-meter patch of land as either a breeding or non-breeding ground.}
    \label{fig:model_prithvi}
\end{figure}

Prithvi \cite{jakubik2023foundation} is a ViT-based geospatial foundational model pre-trained on HLS data. It features a self-supervised encoder with a ViT architecture \cite{dosovitskiy2021image}, incorporating a Masked AutoEncoder (MAE) learning strategy and an MSE loss function. The model exhibits spatial attention across patches and introduces temporal attention. Prithvi demonstrates superior performance in diverse remote sensing temporal tasks, such as multi-temporal cloud gap imputation, floods and wildfire scars segmentation, and multi-temporal crop segmentation \cite{jakubik2023foundation}. Leveraging Prithvi's capabilities, we derive a model that incorporates Prithvi's pre-trained ViT encoder and transpose convolution decoder blocks as illustrated in Figure \ref{fig:model_prithvi}. Our Prithvi-LB model was trained to learn the temporal and spatial intricacies of predicting locust breeding grounds using HLS data.

\section{Experiments}

This section details the training of our models on both categories of input features described in  \ref{sec:input_features}. We first outline our experimental framework and then discuss our experiments.

\subsection{Experimental setup}

We utilized TPU\_V3's for training our ConvLSTM from the Google Cloud platform. Complementing this, our computational setup included 4 vCPUs and 32GB of RAM. For hyperparameter selection, we used a batch size of 32 and trained the models for 200 epochs. Early stopping with a patience of 10 epochs was employed to prevent overfitting. The learning rate was set to $1e-4$, and we utilized the Adam optimizer with $\beta_1$ and $\beta_2$ values of $0.9$ and $0.999$, respectively. 

For the Prithvi-LB model, we used an Nvidia V100 GPU with 8 vCPUs and 30GB of RAM. We fine-tuned the model for 10 epochs with early stopping. The learning rate was set to $1e-4$, and we utilized the AdamW optimizer with $\beta_1$, $\beta_2$ and weight\_decay values of $0.9$, $0.999$ and 0.1 respectively. 

\subsection{Using Remotely-Sensed Input Features}
\label{exp:remote-sensed}
In this experiment, we trained three of our proposed models (PLAN-LB, Conv3D andConvLSTM) and two classical machine learning models—Logistic Regression and Support Vector Machine (SVM) on remotely-sensed input features described in Section \ref{sec:rs_input_features}. Given the spatio-temporal nature of our data, preprocessing for the classical models involved flattening and concatenating the input features. Each of the models was trained to optimize the objective described in \ref{exp:methods} and the best model was selected using the validation split. The performance of the selected model on the test set was evaluated using accuracy, precision, recall, F1 and ROC-AUC scores. The results obtained from each model are presented in Table \ref{table:experiment_results}.


\begin{table*}[t]
\caption{\textit{Experiment Results.} RS and MS refer to remotely-sensed and multi-spectral inputs, respectively. These results were obtained from training different models described in Section \ref{models} using various input types as discussed in Section \ref{sec:input_features}. The table presents performance metrics, including accuracy, F1-score, precision, recall, and ROC-AUC score, for different models in predicting DL breeding grounds. Notably, Prithvi-LB trained using multi-spectral earth observation images yields the highest predictive performance.}
\label{table:experiment_results}
\begin{center}
\resizebox{1.5\columnwidth}{!}{
\begin{tabular}{l|cccccc}
\toprule
\textbf{Methods} & \textbf{Accuracy} & \textbf{F1-score} & \textbf{Precision} & \textbf{Recall} & \textbf{ROC\_AUC} & \textbf{Input} \\
\midrule
SVM & 62.36 & 63.19 & 72.47 & 71.04 & 71.04 & RS \\
Logistic Regression & 60.23 & 60.13 & 69.45 & 67.96 & 67.96 & RS \\
PLAN-LB & 71.21 & 56.94 & 79.84 & 59.20 & 75.09 & RS \\
Conv3D & 75.38 & 64.74 & \textbf{86.29} & 64.67 & 69.91 & RS \\
ConvLSTM & 75.76 & 67.34 & 80.37 & 66.48 & 63.61 & RS \\
\midrule
Prithvi-LB & \textbf{83.03} & \textbf{81.53} & 82.12 & \textbf{82.90} & \textbf{87.69} & MS \\
\bottomrule
\end{tabular}
}
\end{center}
\end{table*}

\subsection{Using Multi-Spectral Earth Observation Images}
\label{exp:multi-spectral}
We trained Prithvi-LB on spatio-temporal chips ($3 \times 6 \times 224 \times 224$) derived from HLS data as described in \ref{sec:hls_input_features}. In this experiment, the model was trained using a segmentation objective, where each pixel has a probability score for the breeding and non-breeding class. The model was trained for 10 epochs and the best checkpoint was selected using the validation split. The results obtained after evaluating the selected checkpoint on the test split are shown in Table \ref{table:experiment_results}.


\section{Results and Discussion}
The results from the experiments conducted in Sections \ref{exp:remote-sensed} and \ref{exp:multi-spectral} are summarized in Table \ref{table:experiment_results}. In the initial set of experiments utilizing remotely-sensed input features, the deep learning models outperformed the classical models, with the ConvLSTM model emerging as the top performer. It achieved an accuracy, F1-score, and ROC-AUC score of 75.76\%, 67.34\%, and 63.61\%, respectively.

On the other hand, the experiments involving HLS multi-spectral earth observation data exhibited the best overall performance, surpassing ConvLSTM with improvements of +5.11, +13.26, and +25.28 in accuracy, F1-score, and ROC-AUC score, respectively. This outstanding performance can be attributed to various factors, including the utilization of a pre-trained Prithvi ViT encoder and the high spatial resolution (30m) of the data. Furthermore, HLS boasts a high update frequency of 2-3 days, enhancing the model's suitability for operational deployment.

It is however noteworthy that a significant number of samples in the breeding records dataset were lost due to missing values on one or more variables during the preprocessing phase for remotely-sensed input features. This issue not only affects the performance and reliability of remotely-sensed input features but is also exacerbated by the low update frequency of the various variable sources, rendering them less suitable for models intended for operational deployment.

\begin{figure}[ht]
\centering
\includegraphics[trim={0 1cm 12cm 0}, clip, width=0.95\linewidth]{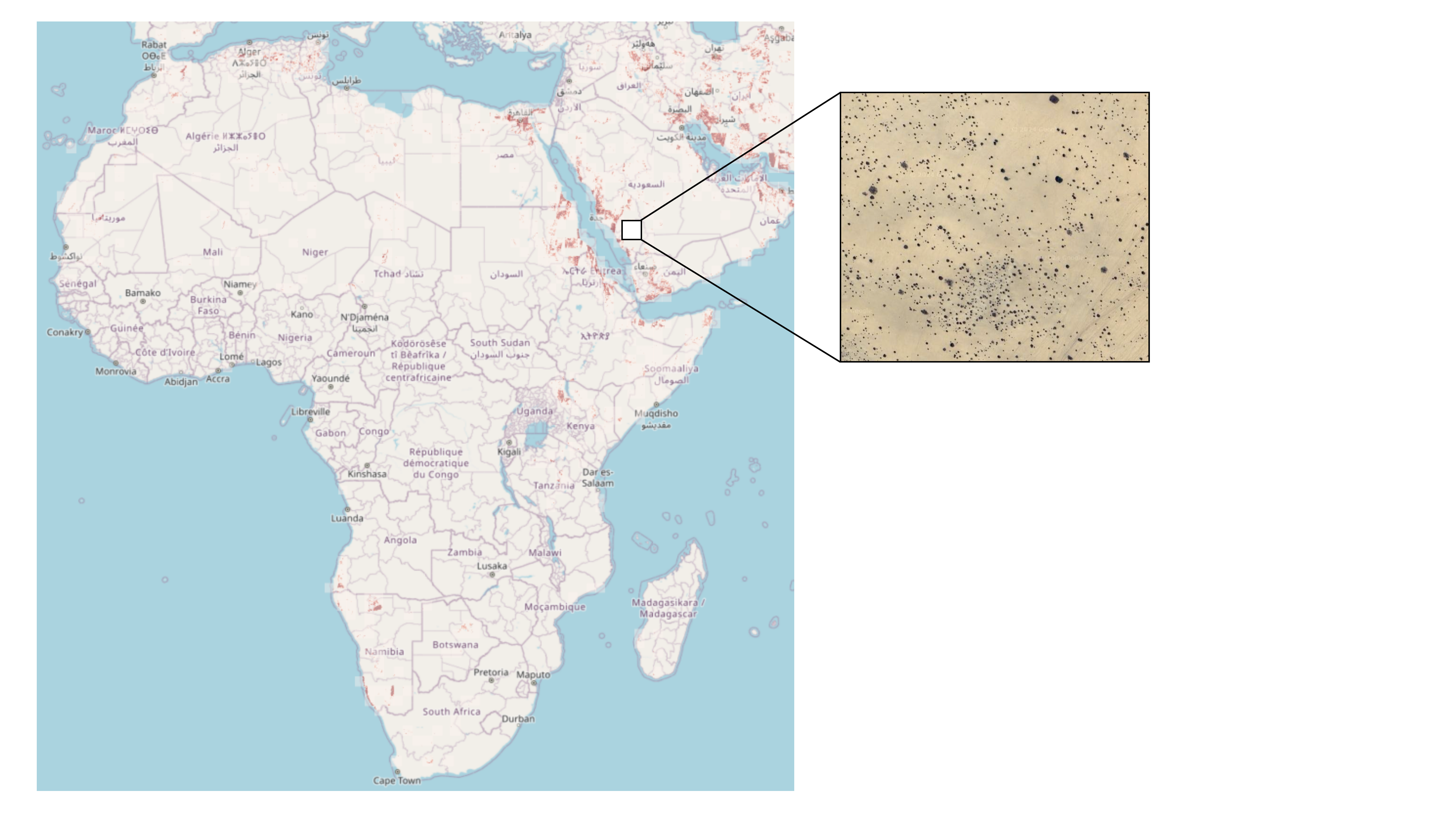}
\caption{\textit{Visualization of Prithvi-LB Predictions for January 2023.} This figure presents the spatial predictions generated by the Prithvi-LB model for January 2023, encompassing every $30m \times 30m$ parcel within the specified study area, as depicted in Figure \ref{breeding_data}. Areas covered in red are predicted as breeding sites. The model adeptly identifies the sparse nature of DL breeding areas. Overlaying these predictions onto satellite imagery reveals that the majority of areas predicted as potential breeding sites are characterized by desert landscapes with sparse tree presence. This pattern suggests these regions might witness vegetation growth following rainfall, indicating possible DL breeding grounds.}
\label{fig:prediction_viz}
\end{figure}

A visual analysis of the predictions made by Prithvi-LB is depicted in Figure \ref{fig:prediction_viz}. It predicts potential DL breeding sites across every $30m \times 30m$ section of our study area, as illustrated in Figure \ref{breeding_data}. The results demonstrate the model's capability to identify the sparse nature of DL breeding grounds effectively. By overlaying these predictions on high-resolution satellite imagery, it was observed that areas identified as probable DL breeding sites predominantly consist of desert terrains with sparse tree cover. These areas are presumed to be where DL edible vegetation might emerge following periods of rainfall. To further substantiate the accuracy of our model's predictions, we plan to conduct ground-based verification in collaboration with the UN-FAO and other partner organizations.

\section{Conclusion}
In this research, we aimed to develop a robust and operationally-ready model for predicting locust breeding grounds, addressing a critical need in managing the threat posed by DL to animal and food security. Utilizing locust observation records from the UN-FAO, along with two categories of input features – remotely-sensed data and multi-spectral earth observation images – we trained and evaluated various models. Our findings indicate that our Prithvi-based model, which utilizes multi-spectral earth observation images, demonstrates superior performance, attaining accuracy, F1 and ROC-AUC scores of 83.03\%, 81.53\% and 87.69\% respectively. This model's effectiveness is largely attributed to leveraging the high temporal (2-3 days) and spatial (30m) resolution Harmonized Landsat and Sentinel-2 (HLS) satellite product. Consequently, our research offers significant advancements in predicting desert locust breeding grounds, with potential for enhancing the administration and effectiveness of control activities undertaken by regional governments and relevant agencies.







\section*{Impact Statements}
Our proposed methodology for detecting DL breeding grounds solely utilizing multi-spectral earth observation images has not only surpassed existing approaches but has also demonstrated the potential for immediate operational deployment. This advancement addresses a critical need for organizations involved in locust control operations, potentially enhancing their ability to effectively mitigate the threat posed by DL. However, the DL threat is a multi-stakeholder problem that needs complex coordination between many partners and will not be solved by modeling alone. There are also implications of relying solely on model predictions for administering locust control activities, and such a strategy might fail to provide enough information to effectively dispatch (often very expense) control measures.


\bibliography{example_paper}
\bibliographystyle{icml2024}

\end{document}